\pdfoutput=1
\documentclass[conference]{IEEEtran}
\IEEEoverridecommandlockouts

\usepackage[T1]{fontenc}
\usepackage[utf8]{inputenc}
\usepackage{cite}
\usepackage{amsmath,amssymb,amsfonts}
\usepackage{graphicx}
\usepackage{booktabs}
\usepackage{multirow}
\usepackage{array}
\usepackage{xcolor}
\usepackage{url}
\usepackage{tikz}
\usetikzlibrary{patterns,arrows.meta,calc,positioning}
\usepackage{pgfplots}
\pgfplotsset{compat=1.18}
\usepackage[hidelinks]{hyperref}
\usepackage{fancyhdr}
\fancypagestyle{preprint}{\fancyhf{}%
  \fancyhead[C]{\footnotesize\itshape Casta\~no Molina, J. and Garc\'ia de Soto, B. - Pano2BIM (preprint)}%
  \fancyfoot[C]{\footnotesize\thepage}}
\graphicspath{{figures/}}

\newcolumntype{L}[1]{>{\raggedright\arraybackslash}p{#1}}

\begin{document}

\title{Pano2BIM: A Geometry-Constrained Vision-Language Agent for As-Built BIM Generation from a Floor Plan and a Single 360\textdegree{} Image}

\author{%
\IEEEauthorblockN{Juan Diego Casta\~no Molina}
\IEEEauthorblockA{S.M.A.R.T. Construction Research Group\\
New York University Abu Dhabi (NYUAD)\\
Division of Engineering\\
Saadiyat Island, United Arab Emirates\\
jdc9881@nyu.edu}
\and
\IEEEauthorblockN{Borja Garc\'ia de Soto}
\IEEEauthorblockA{S.M.A.R.T. Construction Research Group\\
New York University Abu Dhabi (NYUAD)\\
Division of Engineering\\
Saadiyat Island, United Arab Emirates\\
garcia.de.soto@nyu.edu}
}

\maketitle

\begin{abstract}
Digital twins require accurate, up-to-date building information models (BIM) that reflect physical conditions, yet generating and maintaining as-built models remains costly and labor-intensive. Emerging image-based methods can capture visual appearance without inherently producing structured BIM information, while vision-language models (VLMs) struggle with reliable metric object placement. This paper presents \textbf{Pano2BIM}, a geometry-constrained VLM agent that generates approximate IFC4 models at Level of Development (LOD) 200 from a floor plan with known room dimensions and a single 360\textdegree{} photograph. The VLM identifies landmarks and objects, while deterministic tools estimate camera pose, compute metric positions through ray casting, and check spatial constraints. An optional second stage enriches the presentation model through render-and-compare feedback without modifying the primary furniture geometry. In a preliminary evaluation on one laboratory, the proposed Measure workflow achieved a mean furniture-footprint intersection over union (IoU) of 0.53, compared with 0.44 for a single-response baseline using identical modeling instructions, and matched an average of 4.0 versus 2.3 of seven fixed furniture elements across three runs. Centroid disagreements were similar for the four elements consistently identified by both approaches. The measurement stage averaged 3.8 minutes and USD~1.63 in API costs per run, while the complete demonstrated workflow required approximately 42 minutes and USD~11. These findings demonstrate the feasibility of rapid, low-cost as-built BIM generation, with potential applications in maintaining digital twins that reflect changing physical conditions.
\end{abstract}


\begin{figure*}[!t]
  \centering
  \begin{tikzpicture}[font=\footnotesize]
    \node[anchor=south west, inner sep=0] (a) at (0,0) {\includegraphics[width=0.27\textwidth]{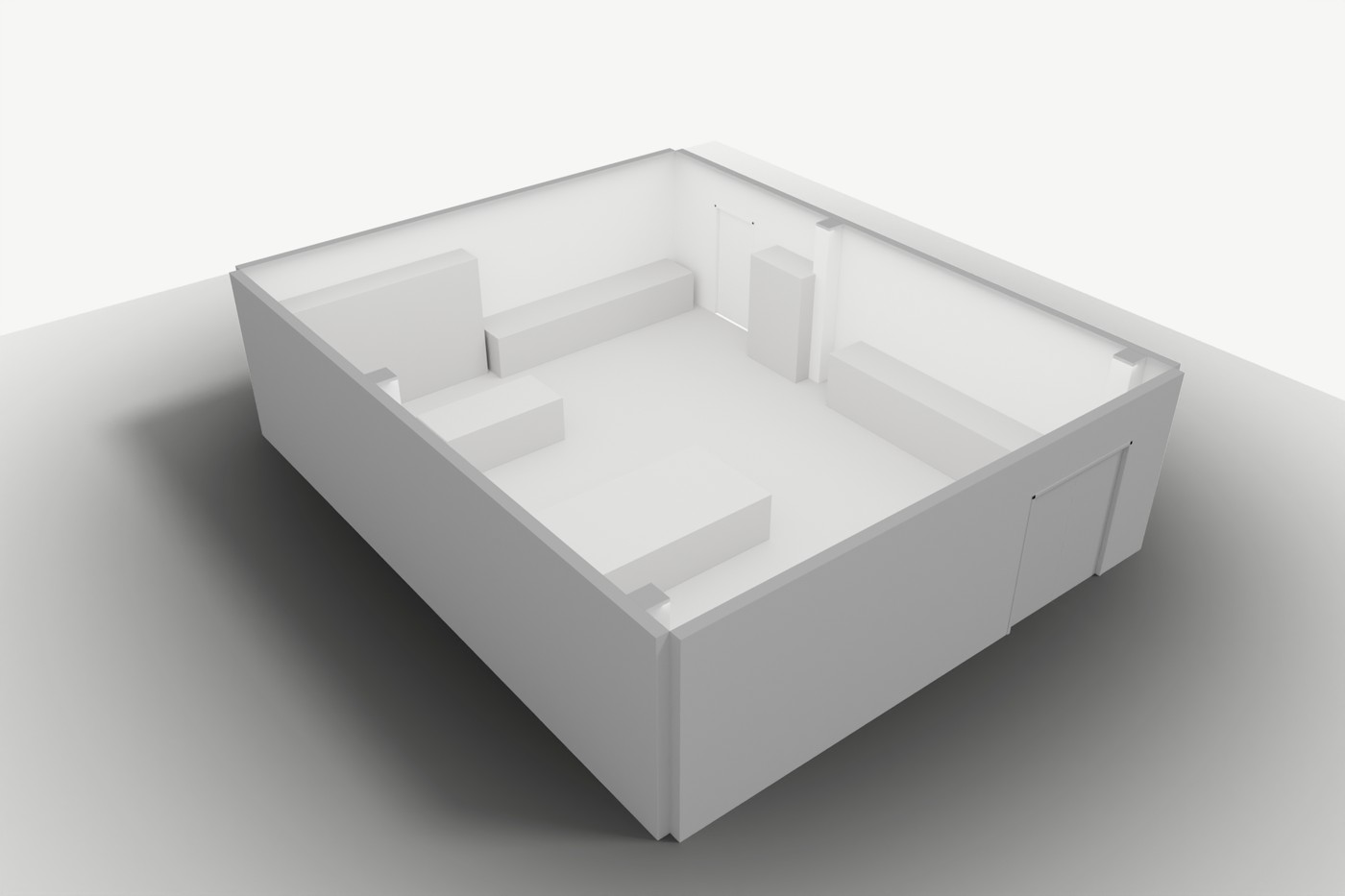}};
    \node[anchor=south west, inner sep=0] (b) at (0.315\textwidth,0) {\includegraphics[width=0.27\textwidth]{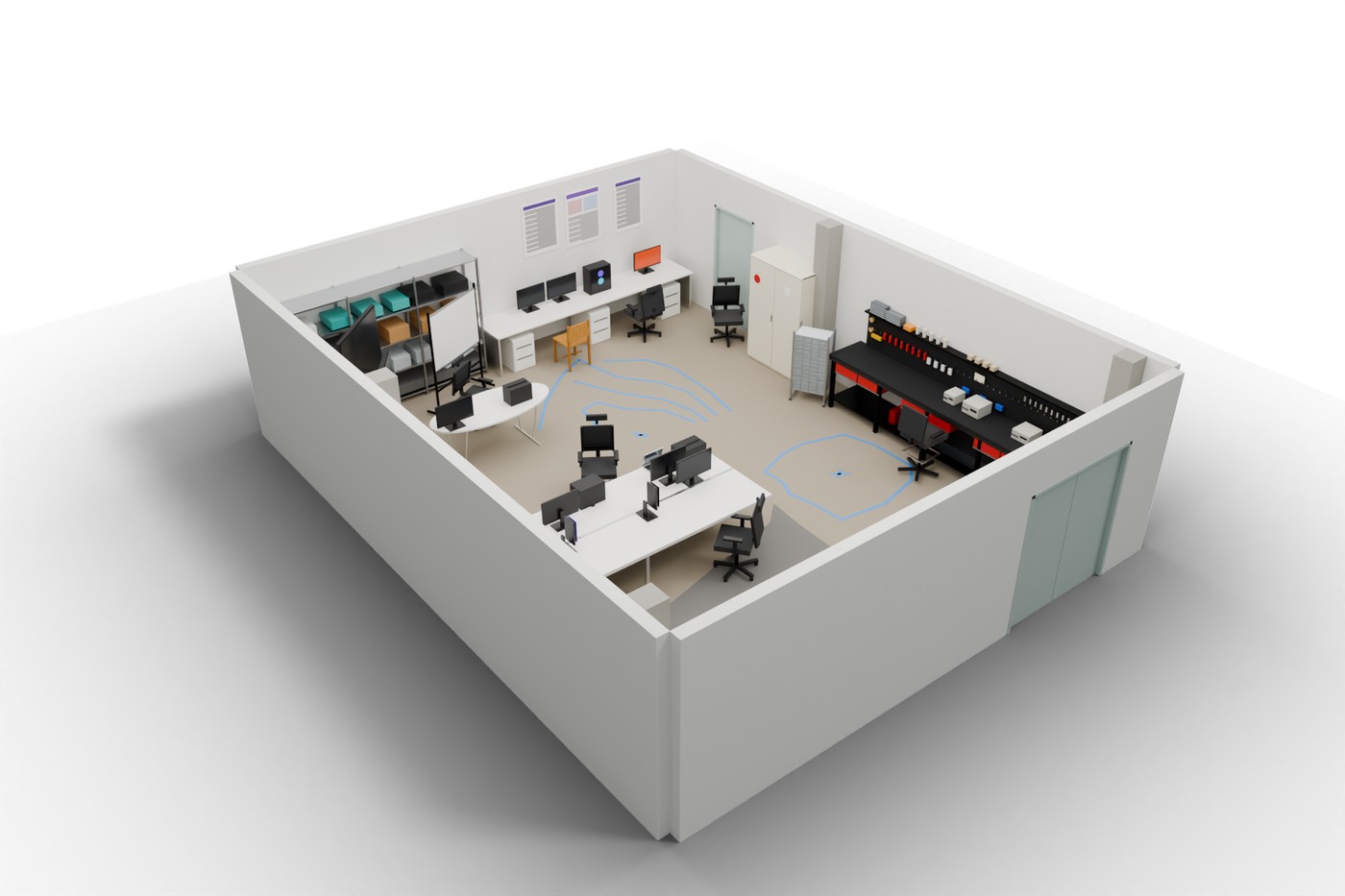}};
    \node[anchor=south west, inner sep=0] (c) at (0.64\textwidth,0) {\includegraphics[width=0.36\textwidth]{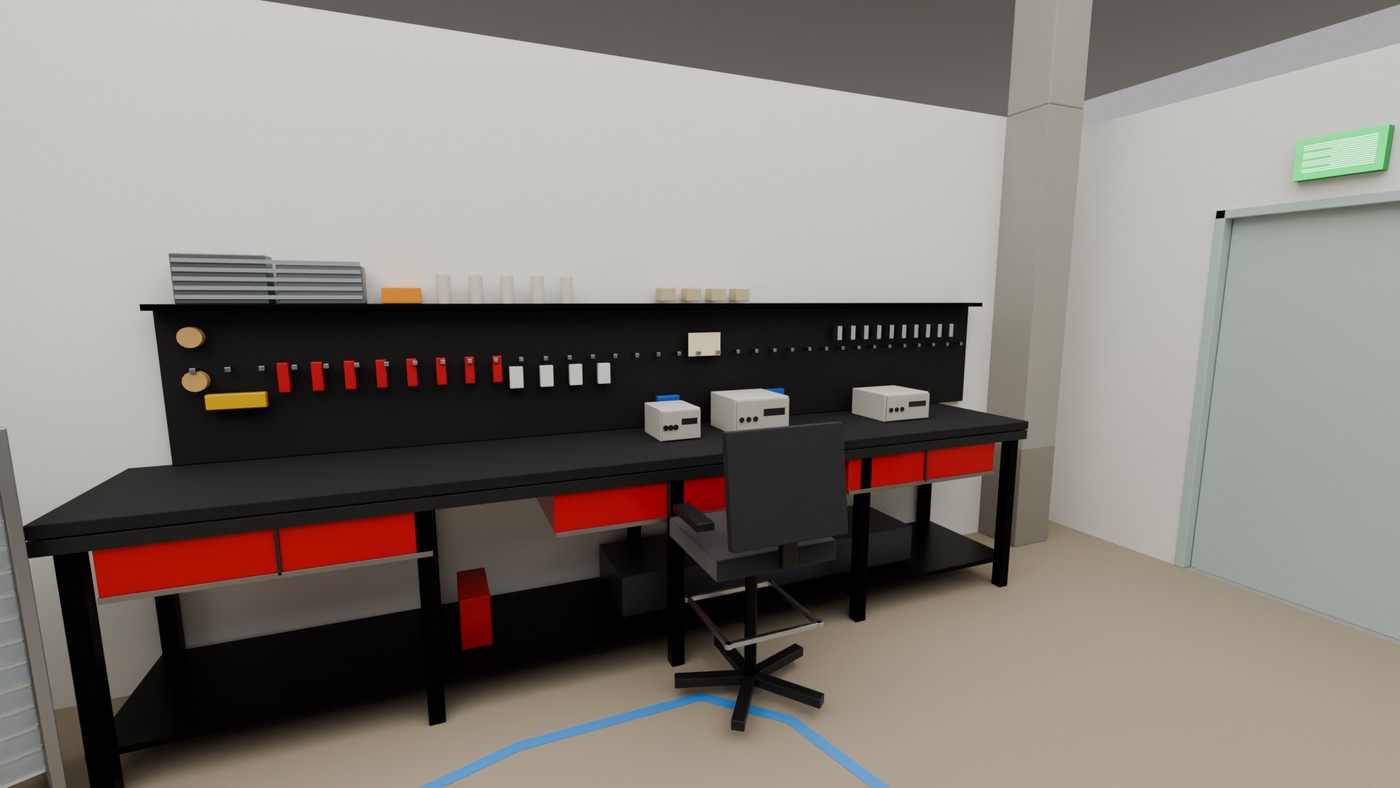}};
    \draw[-{Latex[length=2.2mm,width=2mm]}, line width=1pt, black!70] ([xshift=1mm]a.east) -- ([xshift=-1mm]b.west) node[midway, above, font=\scriptsize] {Stage 2};
    \coordinate (zx) at ($(b.south west)!0.715!(b.south east)$);
    \coordinate (zy) at ($(b.south west)!0.56!(b.north west)$);
    \coordinate (z) at (zx |- zy);
    \draw[line width=1pt, black!70] (z) ellipse [x radius=0.55cm, y radius=0.33cm];
    \draw[densely dotted, line width=0.8pt, black!70] ($(z)+(0.4cm,0.23cm)$) -- ([xshift=-1mm]c.north west);
    \draw[densely dotted, line width=0.8pt, black!70] ($(z)+(0.4cm,-0.23cm)$) -- ([xshift=-1mm]c.south west);
    \node[below=1mm of a, font=\scriptsize] {(a) Measure, 7 boxes (Stage 1)};
    \node[below=1mm of b, font=\scriptsize] {(b) Dress and Verify, same boxes (Stage 2)};
    \node[below=1mm of c, font=\scriptsize] {(c) eye-level view toward the workbench in (b)};
  \end{tikzpicture}
  \caption{Pano2BIM reconstruction of the development space from a floor plan and a single 360\textdegree{} photograph. (a)~Stage 1 (Measure): seven furniture bounding boxes placed using ray casting. (b)~Stage 2 (Dress and Verify): the same bounding boxes enriched with authored shapes, 40 detail objects, and decor. (c)~Eye-level view toward the reconstructed workbench. Stage~2 adds visual detail and leaves the measured furniture geometry unchanged.}
  \label{fig:before_after}
\end{figure*}

\section{Introduction}

Digital twins of existing buildings depend on as-built building information models (BIM) that match each space as it stands today. These models go out of date whenever a room is refitted or its furniture is moved. Facility managers, renovation designers, and construction researchers therefore need reconstruction that is fast and cheap enough to repeat across many rooms and after every change. The standard route is scan-to-BIM, in which a laser scanner or a photogrammetric survey produces a point cloud and a modeler converts it into walls, doors, columns, and furniture~\cite{tang2010automatic,rashdi2022scanning,mahmoud2026indoor}. Both steps are labor-intensive and time-consuming, and the modeling remains largely manual~\cite{rashdi2022scanning,mahmoud2026indoor}, so repeating a survey after every refit is hard to justify for an ordinary room. Neural and Gaussian-splatting reconstructions from video~\cite{kerbl2023gaussian} capture appearance quickly, yet their radiance fields carry no element classes or element dimensions, so a modeler must still convert them into BIM.

A more accessible approach combines an existing floor plan with a single, recently captured 360\textdegree{} photograph. Even when the plan is an undimensioned raster image, known room dimensions allow its geometry, including walls, doors, and columns, to be scaled to metric units, while the photograph captures the furnishings and other visible features as they appear at the time of capture. This paper investigates how a vision-language model (VLM) can interpret these inputs to generate an as-built BIM, with deterministic geometric tools responsible for metric placement. Fig.~\ref{fig:before_after} illustrates the proposed workflow in the development laboratory: Stage~1 (Measure) generates seven furniture bounding boxes (Fig.~\ref{fig:before_after}(a)), while Stage~2 (Dress and Verify) adds visual details without changing their geometric placement (Fig.~\ref{fig:before_after}(b), (c)).

In the experiments of this paper, the VLM identified objects such as workbenches and whiteboards and matched visual landmarks such as doors to the floor plan. When asked to write the whole room model with metric coordinates in one answer, however, it placed furniture less reliably. Given the same modeling instructions as the measuring workflow, the single-call baseline never found the shelving along the north wall and put the workstation block 0.86~m from the expert's position on average (Section~\ref{sec:results}). Benchmarks of large language model (LLM) agents report related limits. No model fully solves more than 3.4\% of IFC editing tasks~\cite{nithyanantham2026bimedit}, procedural 3D modeling fails mostly through API mismatches~\cite{gao2026threedcodebench}, and agentic scene reconstruction is limited by visual precision~\cite{liu2025ir3dbench}. Panoramic scene-understanding networks recover layouts and object boxes from one image~\cite{zhang2021deeppanocontext,dong2024panocontextformer}, but with a fixed vocabulary, a scale learned from training data, and no BIM structure.

Pano2BIM, the workflow presented here, divides the reconstruction between the VLM, which decides what things are, and a set of deterministic geometry tools, which compute where they are. The VLM matches photograph features to plan doors, columns, and corners, names each object and its Industry Foundation Classes (IFC) class, and groups furniture into Level of Development (LOD) 200 elements. It also reads the pixels at each group's ends and edges, authors object shapes and colors, and judges renders against the photograph. The tools compute every position and most extents the photograph reveals. They solve the camera position and heading from the pixel columns of plan landmarks by least squares, estimate its height from points of known height, and intersect each furniture pixel's viewing ray with a plan wall, the floor, or a host top, which gives a point in meters. Other tools later check the dressed model for clashes and unsupported parts. Depths and heights hidden from the camera are VLM estimates from typical dimensions. The VLM never states where an object is in plan coordinates; it gives pixels and a few sizes and offsets, and the tools convert them into positions (Section~\ref{sec:vlmtools}). Every shape it authors is defined relative to a measured box, so a shape can never move an object. A preview loop in Stage~1 and a verify loop in Stage~2 show the agent its model projected into the photograph, and every tool call is typed, logged, and replayable.

The contributions are:
\begin{enumerate}
  \item A low-cost workflow for reconstructing an existing room as BIM, whose measure stage needs only a floor plan with room dimensions and one photograph and is therefore cheap to repeat after the room changes, although updating an existing model is untested. The geometry-constrained agentic workflow runs in two stages, Stage~1 (Measure) and Stage~2 (Dress and Verify), pairs VLM-based semantic interpretation with deterministic plan-constrained placement, and writes IFC4 at LOD~200 with the placement method and a positional uncertainty estimate on every element, which is not yet calibrated (Sections~\ref{sec:placement} and~\ref{sec:limits}).
  \item An agentic harness that turns a 360\textdegree{} panorama into BIM. The VLM agent reads the panorama through a ruled viewer and answers with pixels and labels, and typed tools turn those answers into geometry. In Stage~1 the tools register the camera to plan landmarks, estimate its height, and place each box by casting rays onto a wall, the floor, or a host object (Section~\ref{sec:measure}). In Stage~2, demonstrated in one run, they expand the agent's part-based object types, draw the model's outlines on the photograph, crop matching windows from photograph and render, and report physical faults (Section~\ref{sec:dress}). Every tool call is validated and logged, so each run can be audited (Section~\ref{sec:audit}).
  \item A preliminary case study on one laboratory with one handheld panorama, and an evaluation protocol that scores footprint agreement, element outcomes (matched, placed, or missed), and centroid error against a reference model (Sections~\ref{sec:setup} and~\ref{sec:results}). Over three runs per condition, the Measure workflow reached a footprint intersection over union (IoU) of 0.53 and matched 4.0 of seven elements, against 0.44 and 2.3 for a single-call baseline (Prompt-only) with the same modeling instructions, with similar centroid errors on the elements both found. The dress-and-verify stage is reported as a single demonstration run without a quality measure.
\end{enumerate}
The evaluation uses one space, one photograph, and one VLM; held-out spaces, multi-panorama runs, and independent metric validation are planned (Section~\ref{sec:future}).

\section{Related Work}
\label{sec:related}

\subsection{Scan-to-BIM and image-based modeling}
Automatic as-built BIM from laser-scanned point clouds has been reviewed since Tang et al.~\cite{tang2010automatic}, and recent reviews cover sensors, registration, segmentation, and open BIM standards~\cite{rashdi2022scanning} and the full scan-to-BIM workflow for structured and unstructured elements~\cite{mahmoud2026indoor}. Deep learning segments indoor clouds into object classes and fits library objects to the resulting boxes~\cite{park2022deep}, and knowledge-driven pipelines reconstruct building frames and wall-embedded components whose surfaces lie within 7 to 17~mm (RMSE) of the input point cloud~\cite{xiong2023knowledge}. Related methods compare a scan with a design model~\cite{bosche2010automated} or update parametric elements to match a cloud~\cite{rausch2021automated}. All of them start from a dense metric point cloud.

Image-based methods lower the cost of capture by replacing the scanner with photographs, although the methods below need many overlapping views. Structure-from-motion (SfM)~\cite{schonberger2016structure} on unordered site photographs registered to an IFC model supports progress monitoring~\cite{golparvarfard2015automated}. Mehraban et al.~\cite{mehraban2025automated} label drone or phone frames with open-vocabulary detection and segmentation, recover poses with SfM, and train a neural radiance field (NeRF) to obtain a color-labeled point cloud, from which they fit wall planes and axis-aligned window and door boxes and import them into a BIM authoring tool. The result is the unfurnished building shell, and it requires dense, overlapping video. Wang et al.~\cite{wang2025aei} reconstruct multi-room layouts from a 360\textdegree{} walk-through video by selecting capture points with visual simultaneous localization and mapping (SLAM) and running a panoramic layout network per room, with metric scale from a fixed camera height of 1.9~m. A conference paper by the same group applies a closely related pipeline to 360\textdegree{} video from a building site~\cite{wang2025indoor}. Such methods yield metric room shells from panoramas, although they need a camera trajectory, assume the camera height, and stop at walls, floor, and ceiling. BIM-assisted 360\textdegree{} visual SLAM aligns the trajectory to a design model~\cite{hwang2026toward}, large-scale 360\textdegree{} indoor mapping targets textured geometry~\cite{jung2025im360}, and a survey describes the structured-reconstruction pipeline these systems share~\cite{pintore2020state}. Pano2BIM takes the shell and the metric frame from the plan and uses the photograph for what the plan lacks, the contents of the room.

\subsection{Single-image and panoramic reconstruction}
Single-image methods reconstruct furnished rooms from one photograph or panorama, which brings their capture closest to that of Pano2BIM, although they take scale from learned priors. IM2CAD~\cite{izadinia2017im2cad} estimates the room layout from one photograph, detects objects, retrieves CAD models, and refines their placement by repeatedly rendering the scene and comparing it with the photograph. PanoContext~\cite{zhang2014panocontext} introduced whole-room understanding from a single panorama, and DeepPanoContext~\cite{zhang2021deeppanocontext} and PanoContext-Former~\cite{dong2024panocontextformer} predict layout, object boxes, poses, and shapes from one panorama with networks trained on synthetic and rendered panoramas over a fixed set of object categories. These networks, the closest perception-side alternative to the measure stage, need no plan. Their scale, however, is tied to the training camera height, their evaluation is mainly on synthetic or rendered rooms, and their boxes and meshes carry no BIM classes or provenance. Toderici et al.~\cite{toderici2026objectcentric} generate a textured mesh per detected object in a single interior photograph and place each mesh with monocular metric depth. On 200 synthetic Structured3D scenes, each split into four perspective views, they report a mean placement error of about 1.2~m, which they regard as stable localization, with systematic size errors; their scene-level benchmark sets the kind of multi-scene evaluation this preprint does not yet provide. Panoramic layout estimation~\cite{sun2019horizonnet,jiang2022lgtnet,ibrahem2025panotps}, metric panoramic depth~\cite{lin2025depthanypanoramas,guo2025depthanycamera}, open-vocabulary detection~\cite{liu2023groundingdino}, and promptable segmentation~\cite{kirillov2023segment,ravi2024sam2,carion2025sam3}, surveyed in~\cite{pintore2026state,zhu2026panoramic}, complement these systems.

Pano2BIM applies the analysis-by-synthesis principle~\cite{yuille2006vision,li2018deepim} that IM2CAD uses, with a VLM reading the comparison and an IFC model as output. Because a 1.2~m placement error is large at room scale, Pano2BIM takes its scale and metric frame from the plan and computes placement by ray casting from a camera registered to it. The perception models above are not used in the reported experiments, which rely on the VLM's own pixel readings with the plan as the layout prior.

\subsection{Floor-plan-based reconstruction}
Plan2Scene~\cite{vidanapathirana2021plan2scene} converts a vector floor plan and a few room-assigned photographs into a textured 3D house. The plan gives the architecture, rules add the fixed objects, and the photographs supply one tileable texture per wall, floor, and ceiling, so the method places no furniture, measures nothing in the photographs, and is evaluated on texture quality. Pano2BIM also treats the plan as the fixed architectural frame, and it uses the photograph to register a camera and measure the objects the plan omits, producing a furnished IFC model.

\subsection{Language agents for BIM}
Language agents already author and edit BIM models, mostly from text or existing files. Text2BIM orchestrates several LLM agents that write code against a BIM authoring API and checks the result with rules~\cite{du2025text2bim}. A copilot agent executes modeling tasks inside the authoring tool~\cite{du2024copilot}, and BIMgent drives its graphical interface as a computer-use agent~\cite{deng2025bimgent}. MCP4IFC exposes IFC operations to an LLM through the Model Context Protocol~\cite{nithyanantham2025mcp4ifc}, and BIM-Edit benchmarks 324 IFC editing tasks~\cite{nithyanantham2026bimedit}. On it, the best model averages about 50\% over geometry, semantic, and topology metrics while no model fully solves more than 3.4\% of the tasks. Sketch2BIM converts hand-drawn plans with multimodal agents and a human in the loop~\cite{ratul2025sketch2bim}. Closer to perception, VS-BIM uses panoramic video and BIM information to build a reconstructed scene and a cognitive map for safety inspection~\cite{wang2025vsbim}, and VLMs have been evaluated for hazard identification~\cite{adil2025using}. Pano2BIM starts from a plan and a photograph and produces the measured model itself.

\begin{table*}[t]
  \caption{Among the closest approaches, Pano2BIM integrates plan-registered metric placement of furniture with IFC output and per-element provenance. Methods are compared by inputs, geometry source, output, and validation.}
  \label{tab:compare}
  \centering\footnotesize
  \begin{tabular}{L{0.145\textwidth} L{0.15\textwidth} L{0.19\textwidth} L{0.19\textwidth} L{0.195\textwidth}}
    \toprule
    \textbf{Method} & \textbf{Inputs} & \textbf{Geometry source} & \textbf{Output} & \textbf{Validation} \\
    \midrule
    IM2CAD~\cite{izadinia2017im2cad} & One perspective photo, CAD library & Learned layout and detection; render-and-compare CAD fitting & CAD scene (ShapeNet) & Layout pixel error (Hedau, LSUN); voxel IoU and mAP (SUN RGB-D) \\
    Plan2Scene~\cite{vidanapathirana2021plan2scene} & Vector floor plan, room-assigned photos & Plan extrusion; photos for texture only & Textured house mesh, no furniture & Texture metrics and user study on Rent3D++ \\
    DeepPanoContext~\cite{zhang2021deeppanocontext} & One panorama & Learned layout and pose; relation optimization & Layout, 3D boxes, meshes & Detection mAP, collisions on synthetic iGibson; generalization test on PanoContext \\
    PanoContext-Former~\cite{dong2024panocontextformer} & One panorama & Learned depth prior and transformer & Layout mesh, 3D boxes, meshes & mAP and layout IoU on synthetic and rendered rooms \\
    Mehraban et al.~\cite{mehraban2025automated} & RGB video & SfM and NeRF labeled point cloud & Walls, windows, doors in Revit & Door and window detection; perimeter and footprint IoU on one exterior and one interior case \\
    Wang et al.~\cite{wang2025aei} & 360\textdegree{} walk-through video & Visual SLAM and layout network; assumed camera height & Multi-room layout shell & 3D IoU on 194 test panoramas; area, wall position, length and elevation errors against as-designed BIM on two apartment floor plans \\
    SceneCraft~\cite{hu2024scenecraft} & Text, asset library & LLM constraints, VLM render critique & Blender scene & CLIP score and human rating on synthetic queries \\
    Toderici et al.~\cite{toderici2026objectcentric} & One perspective interior image & Monocular metric depth, image-to-3D per object & Textured meshes (GLB) & Placement error about 1.2~m on 200 synthetic Structured3D scenes \\
    \midrule
    \textbf{Pano2BIM} (this work) & Floor plan with room dimensions, one panorama & Plan-landmark camera registration and ray casting; VLM reads pixels and estimates hidden depths and heights & IFC4 LOD~200 with placement method and uncertainty estimate (not yet calibrated) per element; presentation model; replayable trace & Information Delivery Specification (IDS) conformance; internal consistency; agreement with an expert reference on one real room (no survey yet) \\
    \bottomrule
  \end{tabular}
\end{table*}

\subsection{Agentic 3D scene generation}
SceneCraft~\cite{hu2024scenecraft} is an LLM agent that turns a text description into a Blender scene. It builds a relation graph, optimizes asset positions against scoring functions it writes, revises the result with a vision-language critic that reads renders, and accumulates reusable functions across runs; its evaluation uses text-image similarity and human preference, with no metric scale. Holodeck assembles embodied-AI environments from asset libraries~\cite{yang2024holodeck}. IR3D-Bench asks VLMs to recreate a scene with programming and rendering tools and finds visual precision to be the main limitation~\cite{liu2025ir3dbench}, and 3DCodeBench attributes procedural-modeling failures to API mismatches and floating parts~\cite{gao2026threedcodebench}. Tool-using agents in general interleave reasoning and actions~\cite{yao2023react}, learn to call APIs~\cite{schick2023toolformer}, and improve an output by critiquing it~\cite{madaan2023selfrefine}. Closest in principle, the Geometrically-Constrained Agent (GCA) of Chen et al.~\cite{chen2026geometrically} has a VLM first translate a spatial question into a formal task constraint and then answer it only through deterministic tool calls within that constraint, which separates semantic interpretation from geometric computation for spatial-reasoning benchmarks. Pano2BIM adapts SceneCraft's render-and-revise loop and growing library to an existing room. Its loop compares the model with a photograph, its agent declares shapes relative to measured envelopes, and its output is a metric IFC model with per-element provenance.

\subsection{Differentiation from prior work}
Set against the closest methods by input, geometry source, output, and validation (Table~\ref{tab:compare}), Pano2BIM differs mainly in where its scale and positions come from. Its novelty lies in the integration, demonstrated here in one workflow, of four properties that the compared methods cover only in part. The first is open-vocabulary VLM interpretation of a panorama, and the second is deterministic metric placement constrained by the floor plan, with the camera registered to plan landmarks so that scale comes from the plan. The third is IFC4 output whose property set records the placement method and an uncertainty estimate, not yet calibrated, for every element. The fourth is auditable refinement, in which every correction is a logged tool call that can be replayed. The separation of semantic interpretation from geometric computation follows the same principle as GCA~\cite{chen2026geometrically}; Pano2BIM applies it to the generation of a measured IFC model of an existing room and adds the traceability that the separation makes possible.

\begin{figure*}[t]
  \centering
  \includegraphics[width=\textwidth]{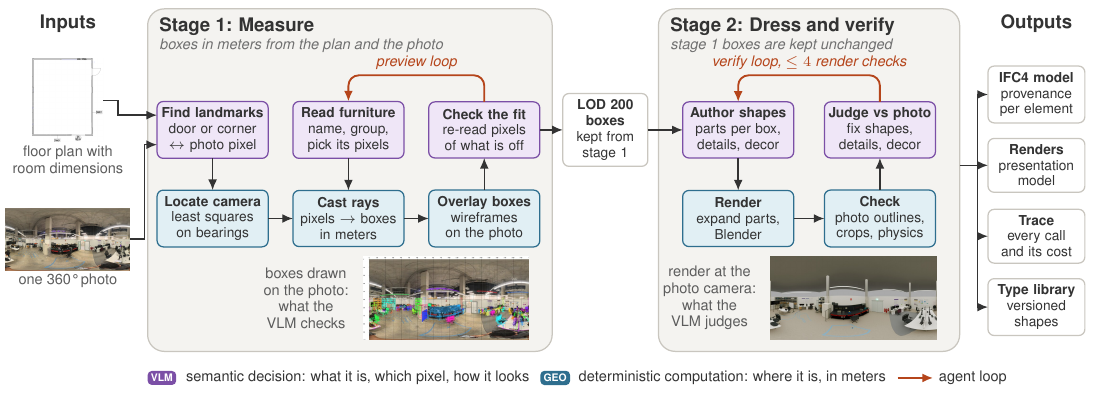}
  \caption{Pano2BIM workflow. Stage~1 (Measure) generates geometrically constrained furniture bounding boxes from a floor plan and a 360\textdegree{} photograph, refined through a preview loop. Stage~2 (Dress and Verify) adds visual details and refines the presentation model through render-and-compare feedback. VLM decisions (purple) guide deterministic geometric operations (blue), producing an IFC4 model, renders, an execution trace, and, optionally, a reusable type library.}
  \label{fig:arch}
\end{figure*}

\section{Method}
\label{sec:method}

Pano2BIM turns a floor plan with the room dimensions and one 360\textdegree{} photo into an IFC4 model in two stages (Fig.~\ref{fig:arch}). Stage~1 (Measure) registers the camera to the plan, that is, finds where the camera stood and which way it faced, and places the furniture as LOD~200 boxes in meters (Section~\ref{sec:measure}). Stage~2 (Dress and Verify) adds detail objects, shapes, colors, and decor, and checks the result against the photo (Section~\ref{sec:dress}). In both stages the VLM works as an agent that looks at the plan and the photo, calls a tool, reads the result, and decides what to do next. A run outputs an IFC4 file, a room model in JSON, renders, and a replayable trace, and a Stage~2 run with the type library enabled also saves the object types it authored (Section~\ref{sec:library}).

\subsection{Inputs and outputs}
\label{sec:inputs}
Each room is stored as a space, which holds a floor plan (DXF, PDF, or raster image), a 360\textdegree{} photo, and a short metadata file. The metadata gives the room's width and length, its wall height and thickness, the direction of north, and the camera height when it is recorded or estimated. When a raster plan has no dimension labels, the room width and length set its scale. The code also accepts several photos per space, a setting this paper leaves untested.

Each run writes a room model, a JSON file that lists the room outline, the camera pose, and every element as a box with its class, size, orientation, and uncertainty estimate. IfcOpenShell~\cite{ifcopenshell} converts it into an IFC4 file~\cite{iso16739}. The room becomes an IfcSpace, while walls, doors, columns, and slabs are assigned their corresponding IFC4 classes. The VLM assigns furniture elements to supported IFC4 classes, using IfcFurniture where applicable and IfcBuildingElementProxy when no suitable specific class is available. Every element carries a property set with its placement method, source photo, and positional uncertainty, and an Information Delivery Specification (IDS) file~\cite{bsi_ids} checks that every element carries this property set and its spatial containment (Section~\ref{sec:levels}).

Following LOD~200~\cite{bimforum_lod}, the model takes the walls, openings, columns, and slab from the plan and represents furniture as approximate boxes, one per functional group: a run of furniture along a wall, a free-standing block used as one unit, or a single item that occupies floor area on its own. Small movable items, items standing on other furniture (such as monitors), and flat items on the walls (such as posters) are left to Stage~2 (Section~\ref{sec:dress}), which adds them as detail objects and decor outside the LOD~200 layer. The prompt states this grouping rule in plain language, and the VLM applies it, including whether a chair counts as a small movable item. The grouping is thus a modeling convention of this study that the evaluation must account for (Section~\ref{sec:reference}).

\subsection{VLM and tools}
\label{sec:vlmtools}
Pano2BIM lets the VLM decide what things are and the tools compute where they are (Table~\ref{tab:origin}, purple and blue in Fig.~\ref{fig:arch}). The VLM never states where an object is in plan coordinates. Its numeric answers are pixel readings, such as the column of a door jamb, and a few quantities that the photo cannot measure, each given relative to the object itself or to a surface the tools have already located: the depth and height of a box, the height above the floor of an outline edge it reads, any gap between a box and its wall, how far across a host an object sits, and, in Stage~2, small offsets that correct a detail object. It estimates these from the photo and typical dimensions. The tools turn all of them into plan coordinates, the camera pose by a least-squares fit and each box by intersecting viewing rays with a wall, the floor, or the top of another object (Section~\ref{sec:measure}). Because the estimates still move a box, its accuracy rests on the pixel readings, the plan dimensions, the registered camera, and the estimated sizes and heights. Section~\ref{sec:levels} defines how these are checked, and Section~\ref{sec:limits} what remains unchecked.

\begin{table}[t]
  \caption{Division of work. The VLM reads pixels and estimates what the camera cannot measure; deterministic tools turn these into every position in meters.}
  \label{tab:origin}
  \centering\footnotesize
  \setlength{\tabcolsep}{3pt}
  \begin{tabular}{L{0.21\columnwidth} L{0.355\columnwidth} L{0.355\columnwidth}}
    \toprule
    \textbf{Quantity} & \textbf{VLM reads or estimates} & \textbf{Tools compute} \\
    \midrule
    Room shell & -- & Taken from the plan and metadata \\
    Camera position and heading & Pixel column of each matched plan landmark & Least-squares fit to the landmark bearings \\
    Camera height & Pixel rows of points with a known height (door foot, door head) & Median over those rows; a tripod record takes precedence; else the photographer's estimate or 1.5~m (flagged) \\
    Box against a wall & Wall, end columns, depth, height & Box ends from rays to the wall line; yaw from the wall \\
    Free-standing box & Outline points (column, row, edge height), height; width and depth when the points do not fix them & Footprint from rays to the plane at each point's edge height; yaw snapped to the room axes within 20\textdegree \\
    Object on a host (desk, shelf) & Host, center column, depth fraction across the host, size & Position along the line of sight, base on the host top \\
    Uncertainty estimate & -- & Grows with distance; set by anchor type (Section~\ref{sec:placement}) \\
    Meaning & Identity, IFC class, grouping, facing & -- \\
    Appearance & Shapes as parts, colors, decor edges in pixels & Parts fitted in the measured box; decor in meters by rays \\
    \bottomrule
  \end{tabular}
\end{table}

\begin{figure*}[t]
  \centering
  \includegraphics[width=\textwidth]{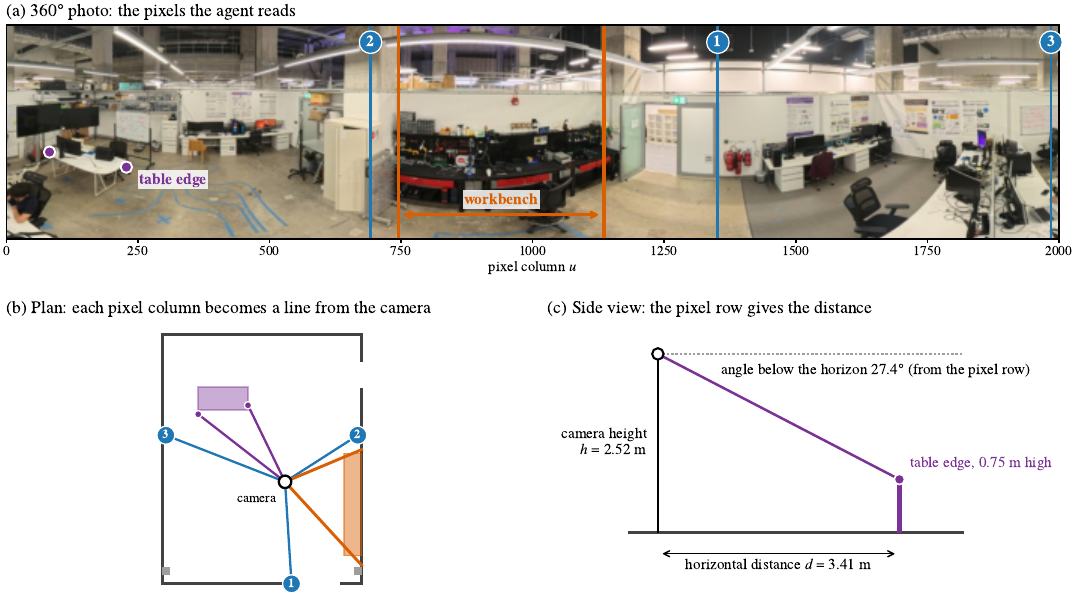}
  \caption{Stage~1 converts pixels into directions and distances from the camera. (a)~Pixels read by the agent in the first Measure run, namely three of its plan landmarks (blue), the two ends of the workbench (orange), and two points on the edge of the oval table (purple). (b)~In the plan, each pixel column becomes a line from the camera. The blue lines fix the camera, and the orange and purple lines place the two boxes. (c)~For one purple point, the angle below the horizon, read from the pixel row, and the camera height together give the horizontal distance.}
  \label{fig:measure}
\end{figure*}

\subsection{Stage 1: measure}
\label{sec:measure}
Stage~1 turns the agent's readings of the 360\textdegree{} photo and the plan into a camera pose and a metric box for each furniture element, and every step relies on one property of the photo that Fig.~\ref{fig:measure} illustrates. In a 360\textdegree{} photo, each pixel column corresponds to a horizontal direction from the camera, and each pixel row to an angle above or below the horizon. After this conversion from pixels to angles, Stage~1 proceeds in three steps. It registers the camera from the columns of landmarks that the plan also shows (blue in Fig.~\ref{fig:measure}), then places furniture along rays from the registered camera (orange and purple), and finally checks the result in a preview loop.

\subsubsection{Pixels and rays}
The agent reads pixel positions through a viewer that labels columns and rows in original pixel coordinates, as on the column axis under Fig.~\ref{fig:measure}(a), and it zooms into windows of the photo when it needs a finer reading. A column can thus be read to about one pixel, which spans $360/2000 = 0.18$\textdegree{} in the 2000-pixel-wide development photo. Because a 360\textdegree{} photo spreads angles evenly across the image (an equirectangular projection), column $u$ and row $v$ of a $W\times H$ image ($H=W/2$) convert directly to the angle $\lambda$ to the right of the image center and the angle $\varphi$ above the horizon:
\begin{equation}
  \lambda(u) = \Big(\tfrac{u+0.5}{W} - \tfrac12\Big)\,2\pi, \qquad
  \varphi(v) = \Big(\tfrac12 - \tfrac{v+0.5}{H}\Big)\,\pi .
  \label{eq:equirect}
\end{equation}
The center column looks along the camera heading $\psi$, measured counterclockwise from east, so a pixel at angle $\lambda$ looks along the plan bearing $\psi - \lambda$. Each colored line in Fig.~\ref{fig:measure}(b) is such a bearing, drawn from the camera. Plan coordinates are in meters from the inner south-west corner ($+x$ east, $+y$ north, $+z$ up).

\subsubsection{Camera registration}
Camera registration finds where the camera stood and which way it faced by matching photo columns to landmarks whose plan positions are known. The agent identifies at least four well-spread plan landmarks in the photo, such as room corners, door jambs, and column centers, checks that they appear from left to right in the clockwise order the plan implies, and reads the column $u_i$ of each. The blue lines 1 to 3 in Fig.~\ref{fig:measure}(a) are three of the six landmarks read in the first Measure run. For a candidate camera position $(x,y)$ and heading $\psi$, each landmark has a predicted bearing, and registration chooses the pose whose predicted bearings agree best with the read ones by minimizing the sum of the squared residuals
\begin{equation}
  r_i(x,y,\psi) = \operatorname{wrap}\!\Big(\psi - \operatorname{atan2}(Y_i - y,\ X_i - x) - \lambda(u_i)\Big),
  \label{eq:residual}
\end{equation}
where $(X_i,Y_i)$ is the plan position of landmark $i$ and $\operatorname{wrap}$ maps angles into $[-\pi, \pi)$; residuals are reported in degrees. At the solved pose, the line through each read column passes through its landmark, as the blue lines in Fig.~\ref{fig:measure}(b) do. A rectangular room looks similar after a half turn, so the bounded least-squares fit~\cite{virtanen2020scipy} starts from 108 guesses (a $3\times3$ grid of positions at one quarter, one half, and three quarters of each room axis, times twelve headings 30\textdegree{} apart), keeps the best, and rejects any solution outside the room polygon. The tool returns the root-mean-square (RMS) residual and the predicted column of every plan landmark, including unused ones, so that the agent can check them in zoomed views. The prompt accepts a pose only when the RMS residual is at most 2\textdegree{} and the predicted columns agree with the photo.

Distances along the floor also require the camera height $h$, which a handheld shot does not record, so registration estimates it from points of known height. The agent reads the row $v_i$ of such points, a door foot ($z=0$) or a door head (the door height from the plan or metadata, else a default of 2.10~m, which the development runs used). Fig.~\ref{fig:measure}(c) shows the same triangle for a table edge, whose height of 0.75~m the agent estimates. A point at horizontal distance $d_i$ from the camera, known once the pose is solved, and seen at angle $\varphi(v_i)$ lies $d_i\tan\varphi(v_i)$ above or below the camera, so each point gives one estimate of $h$, and the median keeps a single bad reading from moving the result:
\begin{equation}
  h = \operatorname{median}_i\big( z_i - d_i \tan\varphi(v_i) \big).
  \label{eq:height}
\end{equation}
Across the three Measure runs on the development photo, registration placed the camera between $(4.86, 4.08)$ and $(4.93, 4.15)$~m, with $h$ between 2.46 and 2.52~m, headings within 0.6\textdegree{} of one another, and RMS residuals of at most 1.1\textdegree.

\subsubsection{Ray-cast placement}
\label{sec:placement}
Once the camera is registered, each pixel column fixes a direction in the plan, and the placement tool finds the distance along that direction by intersecting the ray with a surface of known position, called an anchor. Fig.~\ref{fig:measure} shows two of the three anchors, a wall for the workbench (orange) and the floor for the oval table (purple). Depths and heights that the photo cannot show, such as the depth of the workbench, are the agent's estimates from typical sizes (Table~\ref{tab:origin}).
\begin{itemize}
  \item A wall anchor places a run of furniture against a wall. The agent names the wall and reads the columns of the run's two ends, the orange lines in Fig.~\ref{fig:measure}(a). Each end column gives a horizontal ray, which meets a line parallel to the wall at half the box depth from it, plus any gap between wall and box that the agent states. The two intersections, clipped to the wall length, give the ends of the box, and the wall gives its orientation, as Fig.~\ref{fig:measure}(b) shows. Wall anchors use columns only, so they do not depend on the camera height.
  \item A floor anchor places a free-standing object. The agent reads up to six points on its outline (the prompt asks for two to four), each with its row and the height $z$ of the edge it shows, from $z=0$ at the floor contact to $z=0.75$~m for the table edge in Fig.~\ref{fig:measure}(a). As Fig.~\ref{fig:measure}(c) shows, the row gives the angle below the horizon, and the camera height turns that angle into the horizontal distance
  \begin{equation}
    d = \frac{h - z}{\tan\!\big(-\varphi(v)\big)}
    \label{eq:floor}
  \end{equation}
  along the point's bearing. One point plus a width gives a box facing the camera, two points give the front edge and its orientation, as for the table, and three or more give a minimum-area rectangle. An orientation within 20\textdegree{} of a room axis snaps to it. The tool reports the distance of every point, so that points read in the nadir region, where the photographer occupies the image, can be spotted. Points far from the camera sit close to the horizon, where a one-pixel error in the row moves the point far, so the prompt asks for the nearest corners.
  \item A host anchor places an object on top of another box, mostly a Stage~2 detail object. Its points meet the plane of the host's top or, when its base is hidden, the agent gives a fraction across the host along the line of sight. The tool keeps the object inside the host footprint.
\end{itemize}

Each furniture element is assigned a heuristic positional uncertainty indicator based on its distance from the camera and placement method. This indicator reflects expected sensitivity to image-reading and geometric assumptions but is not a statistically calibrated estimate of reconstruction accuracy. The tool records this estimate, $\sigma_{xy}$, per box axis, and the IFC export writes the root sum of squares of the two axis values to the property set as one number (PosSigma\_m). It grows with the length of the rays in Fig.~\ref{fig:measure}(b), because a fixed reading error moves a far point more than a near one. Along a wall the estimate is 2.6~cm per meter of distance from the camera (an angle of 1.5\textdegree), at least 5~cm; the angle leaves a margin over a one-pixel reading error (0.18\textdegree) plus a typical pose residual of about 1\textdegree. Across the wall it is 10~cm, since the depth is estimated. Floor anchors get 6~cm per meter of mean point distance in both directions, at least 8~cm, because their points also depend on the row angle of Fig.~\ref{fig:measure}(c). At 5~m, the estimate is about 13~cm along a wall (16~cm combined with the 10~cm across it) and 30~cm in each direction on the floor (42~cm combined). A survey has yet to calibrate these values.

\subsubsection{Preview loop}
\begin{figure*}[t]
  \centering
  \includegraphics[width=0.8\textwidth]{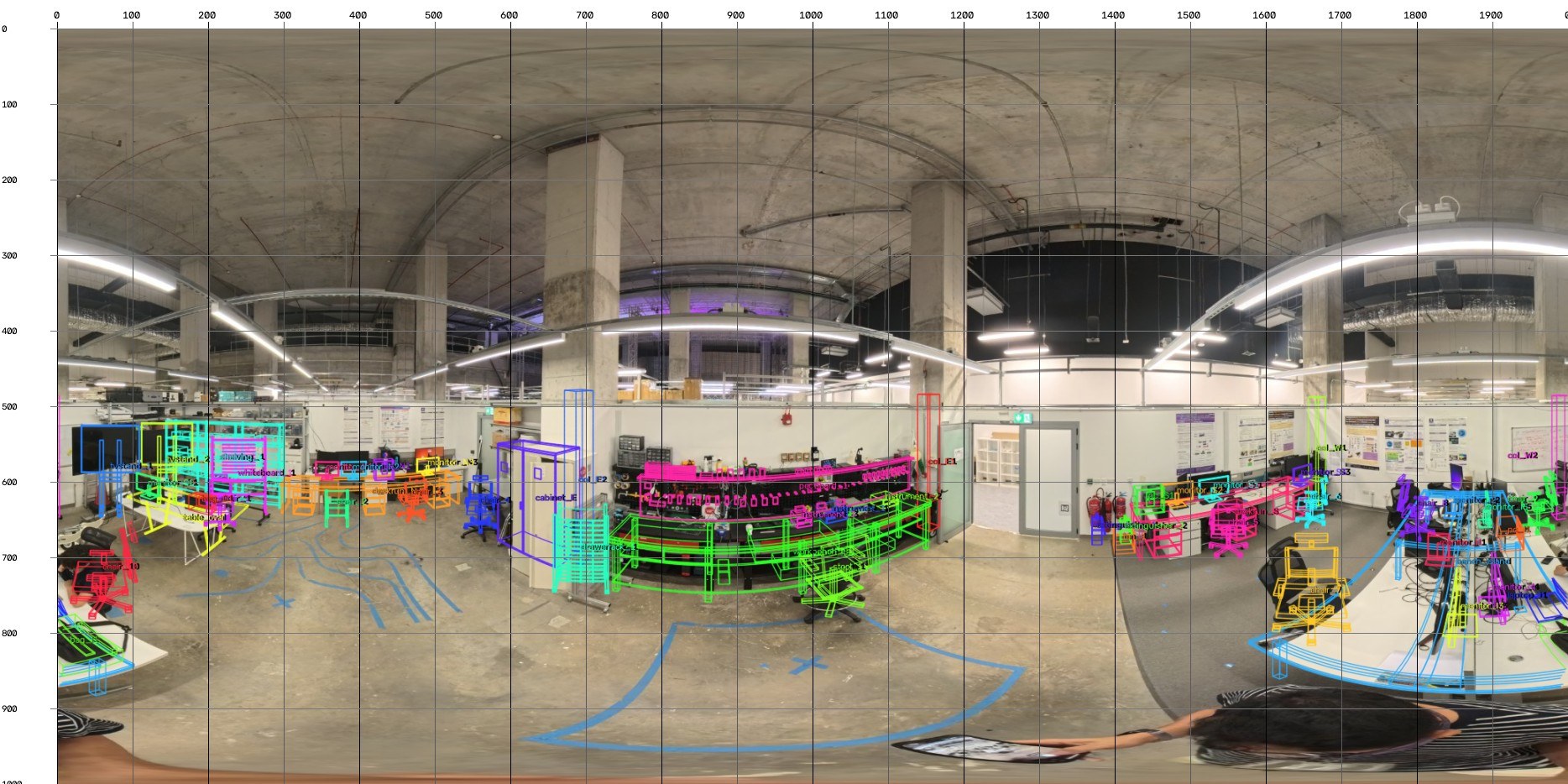}
  \caption{Outlines on the photo, the first Stage~2 check. Wireframes of every element, drawn on the 360\textdegree{} photo from the solved camera, show the agent which objects sit off their place, are missing, or are extra.}
  \label{fig:outlines}
\end{figure*}

The preview loop lets the agent check its placements in the photo, where errors are easier to see than in coordinates. The preview tool returns a plan view of every box on a 1~m grid, similar to Fig.~\ref{fig:measure}(b), and the photo with every element projected back as a wireframe, similar to the overlay of Fig.~\ref{fig:outlines}. The agent checks that each wireframe fits its object and that no box crosses a wall, overlaps another box, or blocks a door. It then re-reads the pixels of a misplaced box and places it again under the same id, or removes it, until another round would change nothing visible; the prompt asks for at least two previews, and a run may use six. The stage ends with IFC export and a summary of what was modeled, what was left out, and which boxes are least certain.

\subsection{Stage 2: dress and verify}
\label{sec:dress}
Stage~2 turns the measured room model into a presentation model that looks like the photo, and this paper demonstrates it once (Section~\ref{sec:dressresults}). The agent adds detail objects, shapes, colors, and decor and checks the result against the photo in rounds. The LOD~200 layer stays as Stage~1 measured it, since the Stage~2 correction tool, which moves detail objects by small offsets in meters, accepts only detail objects, and the prompt forbids moving or removing the measured boxes.

\subsubsection{Types and decor}
A type describes how one kind of object looks, as a list of parts or as a reference to a public 3D model~\cite{polyhaven}. A part is a box, cylinder, sphere, or cone with a center and size given relative to the element's box, a rotation, a color, a finish, and an optional texture, and a repeat copies a part along one axis, as for legs or shelves. The tool scales the parts to each element's measured box and reports any part that leaves it by more than 2~cm, so a type sets only how an object looks and the box sets where it is. Detail objects, the smaller items that the LOD~200 layer leaves out, are placed with the Stage~1 placement tool (Section~\ref{sec:measure}). Decor covers everything flat on the walls or the floor. Posters, signs, and boards are located by the pixel columns and rows of their edges and drawn as colored blocks with any legible text, and floor markings are traced as lines or filled outlines through floor pixels, so no pixels are copied from the photo onto the model.

\subsubsection{Checks}
After each dressing round, three tools let the agent compare its model with the photo. Outlines on the photo draws wireframes of all parts onto the 360\textdegree{} photo from the solved camera (Fig.~\ref{fig:outlines}) and compares the agent's count of each kind of object with the model. Photo-versus-render crops cuts the same window from the photo and from a render of the model, so that shapes and colors can be compared up close. The render check renders a labeled top view, an axonometric view, and a 360\textdegree{} view in Blender Cycles~\cite{blender} and adds a physics report. The report lists hard issues that the agent must fix: a detail object that leaves the room polygon by more than 2~cm or overlaps a column by more than 0.01~m$^2$, two parts of different elements that share more than 0.5~L of volume (parts resting on each other excluded), and a detail object that is not wall-mounted, more than 5~cm above the floor with no part under its center whose top lies within 12~cm of its base. The prompt asks for at least two rounds of checks (the verify loop of Fig.~\ref{fig:arch}), and a run may use four render checks and eight crop inspections of up to eight objects each. The first attempt to finish is refused while a type in use is uninspected and inspection budget remains.

\subsection{Levels of checking}
\label{sec:levels}
Together, the checks of both stages test the model at three levels, and passing one level says nothing about the next.
\begin{enumerate}
  \item Conformance means that the IFC file passes the IFC4 schema and an IDS that requires the property set and the containment relations on every element. The file is well formed and traceable, and its correctness is still open.
  \item Internal consistency means that the model agrees with itself, with the plan, and with the photo seen from the solved camera. Besides the pose residuals, overlays, and physics report, three rules test the exported boxes. The clash rule allows no box to overlap a wall, a column, or another box by more than 2~cm. The door-swing rule keeps every box out of each door's swing zone. The plausibility rule requires every furniture box that is not wall-mounted to stand on the floor or on another box inside the room. A consistent model can still be wrong, for example when a wrong camera height scales every floor box.
  \item Agreement with the physical room needs measurements independent of the photo, such as a laser scan. This paper reports agreement with an expert's model (Section~\ref{sec:setup}).
\end{enumerate}

\subsection{Type library}
\label{sec:library}
The type library lets later runs reuse the types that earlier runs authored. When it is enabled, Stage~2 saves every type in use when the agent finishes. Each type is stored as append-only versions with a timestamp, its provenance (space, run, model, and parent type), and a thumbnail. A lookup ranks types by the words their id, category, description, and part names share with the query, and returns the top hits with thumbnails for the agent to copy and adapt. Whether this reuse helps is left to future work (Section~\ref{sec:future}).

\subsection{Auditability}
\label{sec:audit}
Every run can be checked after the fact, because each tool call is validated and logged. Tool inputs must match schemas that forbid unknown fields, and an invalid input returns a readable error to the agent. The trace stores every call with its full arguments, tokens, and cost, and a replay script rebuilds a run without the model. A cost logger prices every model turn and stops a run at a cap (USD~6 by default and USD~20 for Stage~2), and the agent loop stops after 60 turns.

Three further rules keep the runs honest. The IFC class of each element is selected by the VLM from a predefined set of supported IFC4 classes. The tool schema rejects unsupported classifications, and subsequent modeling stages preserve the assigned class. The expert reference sits in a folder that only the evaluator reads, so the agent never sees the model it is scored against. The code and prompts contain no values specific to a room, an object, or an asset, and wall and corner names come from the plan polygon, so a non-rectangular room would get the same tools, although only the rectangular development space has been run.

\section{Experimental Setup}
\label{sec:setup}

\subsection{Development space}
\begin{figure}[t]
  \centering
  \includegraphics[height=1.1in]{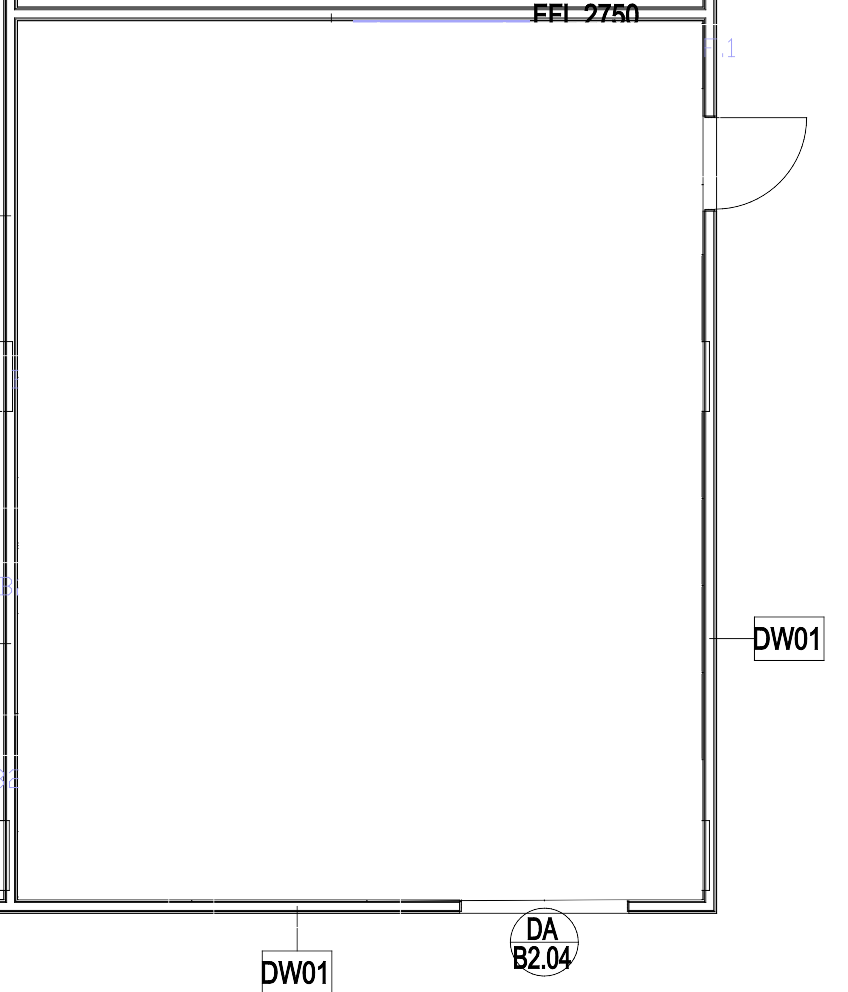}\hfill
  \includegraphics[height=1.1in]{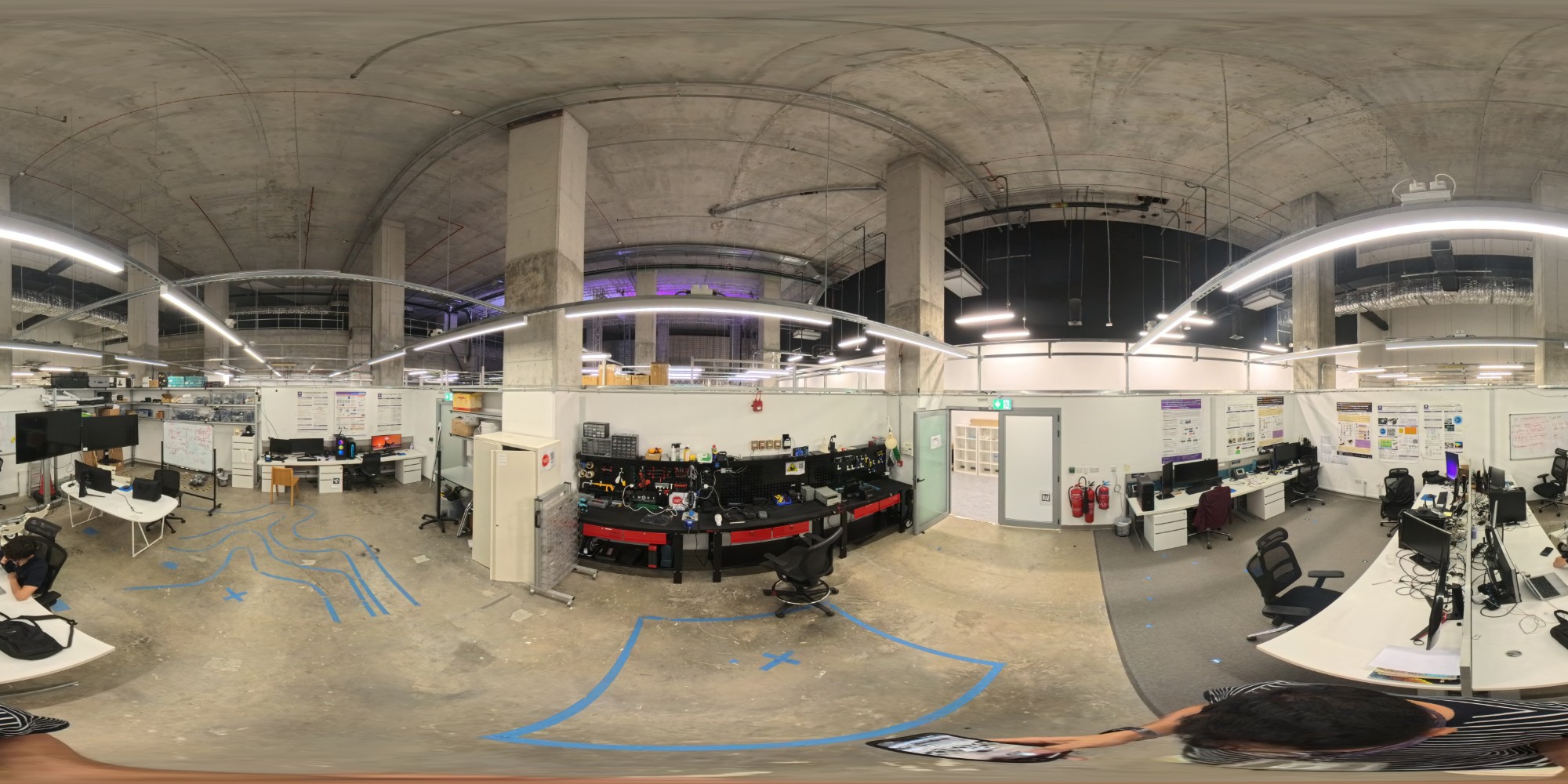}
\caption{The two inputs for the $8\times10$~m development space: the raster floor plan, scaled by the supplied room dimensions (left), and the single $2000\times1000$-pixel 360\textdegree{} photo from a handheld camera (right).}
  \label{fig:inputs}
\end{figure}

The development space, used for every run, is the laboratory of the S.M.A.R.T. Construction Research Group at New York University Abu Dhabi (Fig.~\ref{fig:inputs}). It measures $8\times10$~m and is enclosed by 2.75~m partition walls (from the plan annotation) that stand under a taller hall ceiling, with a double door on the south wall, a single door on the east wall, and columns on the east and west walls. The plan is a raster crop without dimension labels, so the room width and length set its scale. It holds an electronics workbench, cabinets, desk runs, shelving, an oval table, a workstation island, a mobile whiteboard, chairs, monitors, posters, and blue floor markings. The only image is one handheld 360\textdegree{} photo of $2000\times1000$ pixels from a consumer camera. The photographer raised the camera to roughly 2.4~m by eye, and this height was never measured, so Stage~1 estimates the camera height from the photo itself (Section~\ref{sec:measure}).

\begin{figure*}[t]
  \centering
  \includegraphics[width=0.86\textwidth]{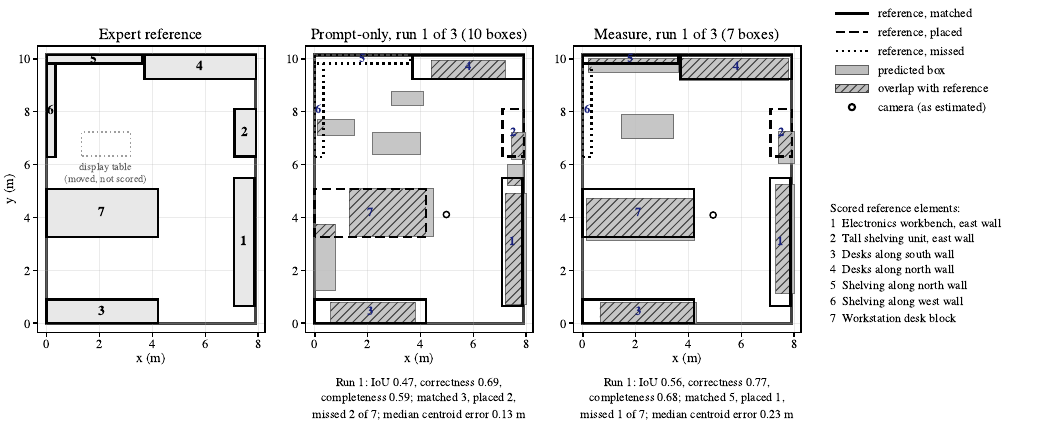}
  \caption{Plan-view comparison of the expert reference (left), Prompt-only baseline (center), and Stage~1 (Measure, right). The first run of each condition, selected before scoring, achieved the highest footprint IoU. Measure matched five of seven fixed reference elements, compared with three for Prompt-only. Metrics shown are for the displayed runs; Table~III reports three-run averages. The movable oval table is shown but excluded from evaluation.}
  \label{fig:plan}
\end{figure*}

\subsection{Reference model}
\label{sec:reference}
The runs are scored against an expert's Revit model of the laboratory, exported to IFC (IFC2X3) in June 2026, four months before the runs. It models every item separately, including eight desks, five shelving units, three workbench segments, an oval table, and twelve chairs. Three steps, fixed before any run was scored, reduce these items to seven LOD~200 elements.
\begin{enumerate}
  \item Align: the model is shifted, with no scaling or rotation, so that the inner south-west corner of its walls is the workflow's origin. Its room measures $7.91\times10.14$~m, slightly different from the $8.0\times10.0$~m that the plan supplies, so the plan's north wall lies 0.14~m south of the reference's and its east wall 0.09~m east of it.
  \item Group: following the rule the agent receives, items of one family that touch (within 5~cm) and run in the same direction form one element, whose footprint is the smallest rectangle around its items. Chairs form no element, because the instructions leave small movable items out of the LOD~200 layer.
  \item Select: the furniture layout changed between the photo and the reference, so only fixed elements, assumed to have stayed in place, are scored. The one movable element, the oval table, is left out, and the mobile whiteboard in the photo is absent from the reference.
\end{enumerate}

The seven scored elements are the electronics workbench and a tall shelving unit on the east wall, the desk rows along the south and north walls, the shelving along the north and west walls, and a workstation block of four desks (Fig.~\ref{fig:plan}, left). Because the reference is one person's model, the numbers in Section~\ref{sec:results} measure agreement with an expert. A survey made at the time of capture, the first item of the planned evaluation (Section~\ref{sec:future}), would turn them into measures of accuracy.

\subsection{Experimental conditions}
Two experimental conditions, each repeated three times on the same floor plan and photograph, use identical furniture-modeling instructions. The Prompt-only baseline generates the complete model in a single response, whereas Measure uses geometric tools, additional image inspections, and iterative refinement. The comparison therefore evaluates the overall geometry-constrained workflow rather than isolating the contribution of deterministic placement alone.
\begin{itemize}
  \item \textbf{Prompt-only} is a single model call with no tools. Given the plan, the room dimensions, the photo, and the format of the room model, it returns the whole room model as JSON with metric coordinates. Its modeling instructions are the LOD~200 grouping rule of the Measure prompt (Section~\ref{sec:inputs}), copied word for word, including what to leave out.
  \item \textbf{Measure} is Stage~1 with the tools of Section~\ref{sec:measure}. It also sees more of the photo, making 12 or 13 model calls and requesting 6 to 12 zoomed views through its ruled viewer besides its previews, whereas Prompt-only answers from the whole photo in one call (Section~\ref{sec:limits} returns to this difference).
\end{itemize}
All runs use Claude Opus 5.5 through the Anthropic API, with adaptive thinking at high effort. The model accepts no sampling temperature, so the runs of a condition differ only through the model's own sampling; all of them reuse the same photo, camera position, and lighting. The cost logger prices input, cached input, cache-write, and output tokens at USD~4.00, 0.20, 5.00, and 20.00 per million, the list prices in October 2026. Stage~2 was run once, on the first Measure run (chosen before scoring), with the type library switched off, so that no type was looked up or saved, and with a per-run cap of USD~20. Its render checks run Blender~5.1 Cycles in background mode from a Python script on a two-core cloud machine.

\subsection{Agreement metrics}
\label{sec:metrics}
\begin{figure}[t]
  \centering
  \includegraphics[width=\columnwidth]{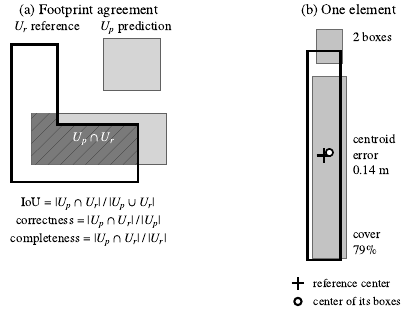}
  \caption{Agreement is measured first on the whole floor and then element by element. (a)~Footprint agreement compares the scored reference footprints $U_r$ (outline) with all predicted footprints $U_p$ (gray) through their overlap (striped). (b)~Element agreement for the workbench (outline) in the first Prompt-only run. Its two assigned boxes (gray) cover 79\% of it, and their center (circle) lies 0.14~m from its center (cross), so it is matched.}
  \label{fig:metrics}
\end{figure}

Agreement with the reference is scored on furniture footprints in plan view through three questions, each illustrated in Fig.~\ref{fig:metrics}. They ask whether a run covers the same floor as the reference, whether it finds each reference element, and how far the elements it finds lie from their reference positions. Table~\ref{tab:results} reports each measure as the mean and range over the three runs of a condition, together with cost (the logger's USD total per run) and wall-clock time.

\begin{table*}[t]
  \caption{Placement agreement with the expert reference (mean and range over three runs).}
  \label{tab:results}
  \centering\footnotesize
  \setlength{\tabcolsep}{3.2pt}
  \resizebox{\textwidth}{!}{%
  \begin{tabular}{l r c c c c c c c r r}
    \toprule
    & & \multicolumn{3}{c}{Footprint agreement} & \multicolumn{3}{c}{Element outcomes (of 7)} & & & \\
    \cmidrule(lr){3-5}\cmidrule(lr){6-8}
    Condition & Boxes & IoU $\uparrow$ & Correctness $\uparrow$ & Completeness $\uparrow$ & Matched $\uparrow$ & Placed & Missed $\downarrow$ & Median centroid error (m) $\downarrow$ & USD & Time (min) \\
    \midrule
    Prompt-only & 9.7 & 0.44 (0.42--0.47) & 0.67 (0.63--0.69) & 0.56 (0.54--0.59) & 2.3 (2--3) & 2.7 (2--3) & 2.0 (2--2) & 0.29 (0.13--0.39) & 0.22 & 1.9 \\
    Measure & 7.0 & 0.53 (0.49--0.56) & 0.75 (0.72--0.77) & 0.64 (0.58--0.68) & 4.0 (3--5) & 1.7 (1--3) & 1.3 (1--2) & 0.31 (0.22--0.50) & 1.63 & 3.8 \\
    \bottomrule
    \multicolumn{11}{l}{Both conditions use Claude Opus 5.5 and the same modeling instructions. $\uparrow$ higher is better; $\downarrow$ lower is better. Outcomes count the seven fixed reference elements; the median centroid error is}\\
    \multicolumn{11}{l}{taken over the four elements that every run found. Model calls per run: one for Prompt-only and 12.7 on average for Measure.}
  \end{tabular}}
\end{table*}

Footprint agreement asks how well the furniture a run draws overlaps the furniture in the reference, regardless of how either is divided into boxes. In Fig.~\ref{fig:metrics}(a), the outline $U_r$ is the union of the seven scored reference footprints, the gray area $U_p$ is the union of all predicted furniture boxes, and the striped area is their overlap. Three ratios divide this overlap by different areas:
\begin{equation}
  \begin{gathered}
  \mathrm{IoU} = \frac{|U_p\cap U_r|}{|U_p\cup U_r|}, \\[2pt]
  \mathrm{correctness} = \frac{|U_p\cap U_r|}{|U_p|}, \qquad
  \mathrm{completeness} = \frac{|U_p\cap U_r|}{|U_r|}.
  \end{gathered}
  \label{eq:footprint}
\end{equation}
Correctness is the striped portion of the gray area, so it falls when a run draws furniture where the reference has none, like the separate gray square. Completeness is the striped portion of the outline, so it falls when a run leaves reference furniture uncovered, like the upper arm of the outline. IoU combines both errors. Correctness and completeness follow the measures of the same names in the ISPRS indoor modeling benchmark~\cite{khoshelham2018evaluation}, computed here on plan footprints with no tolerance buffer. Because all three ratios work on unions, splitting a run of furniture into several boxes changes none of them.

Element outcomes ask whether a run found each of the seven scored elements, which requires deciding which predicted boxes belong to which element. Each predicted box is assigned to the scored element it overlaps most, provided that at least half of the box lies within 0.1~m of that element or the box covers at least half of the element; any other box stays unassigned. The rule uses positions alone, so a box counts for the element it sits on whatever its label says. In Fig.~\ref{fig:metrics}(b), two gray boxes are assigned to the workbench, a 4.2~m workbench run and a cabinet that the model placed against its northern end.

Two numbers then describe each element that received boxes, and together they decide its outcome. Its cover is the share of its footprint that its boxes cover together, and its centroid error is the distance from its center, the cross in Fig.~\ref{fig:metrics}(b), to the center of the union of its boxes, the circle. The workbench has a cover of 79\% and a centroid error of 0.14~m. An element is found when it receives at least one box. A found element is matched when its cover is at least 0.5 and its centroid error at most 0.5~m and placed when it misses either threshold, and an element with no boxes is missed, so the workbench is matched. Center distance separates an object's position from its size and orientation, which is why the nuScenes benchmark matches objects by it~\cite{caesar2020nuscenes}. Taking the center of the union scores an element drawn as several boxes, like the workbench run and the cabinet, by where those boxes lie together.

To compare centroid errors fairly, each run's median is taken over the same four elements, the common set of elements that every run of both conditions found: the workbench, both desk rows, and the workstation block. A median over each run's own found elements would favor a run that skips the hard elements, since a skipped element contributes no error. The common set is determined from the runs themselves, so the medians are conditional on both conditions finding these four elements. Missed elements are counted in the outcomes instead, and Table~\ref{tab:perref} gives the outcome of every scored element in every run and its mean centroid error.

To rule out errors in the scoring code, a second implementation recomputed every score by drawing all footprints on a 1~cm grid, without sharing any geometry code with the first. Apart from two borderline cases where a share lies within 0.002 of 50\%, the two agree within 0.005 on every ratio and cover and within 0.01~m on every centroid error. The north shelving in the second Measure run is matched with a cover of 0.500 (0.498 on the grid), and in the third Prompt-only run the grid assigns one more box to the workstation block, which stays placed. One further outcome lies near a threshold, the south desk row in the third Prompt-only run, which is placed with a centroid error of 0.505~m.

\section{Results}
\label{sec:results}

This section answers the three questions of Section~\ref{sec:metrics} for the two placement conditions, using the summary in Table~\ref{tab:results}, the element-by-element outcomes in Table~\ref{tab:perref}, and the plan views of Fig.~\ref{fig:plan}, and then reports the single dress-and-verify run.

\subsection{Placement}
\label{sec:placeresults}
On the first question, Measure covered the reference furniture more closely than Prompt-only, with a footprint IoU of 0.53 (range 0.49--0.56) against 0.44 (0.42--0.47) (Table~\ref{tab:results}). Both ratios behind the IoU improved by about 0.08: correctness rose from 0.67 to 0.75, so fewer of the Measure boxes fell on empty floor, and completeness rose from 0.56 to 0.64, so less of the reference furniture was left uncovered.

On the second question, Measure matched 4.0 of the seven fixed elements per run on average and missed 1.3, whereas Prompt-only matched 2.3 and missed 2.0. The per-run counts were 5, 3, and 4 for Measure and 3, 2, and 2 for Prompt-only, so the ranges meet at three. Two elements account for the whole difference in matches (Table~\ref{tab:perref}). Only Measure found the north shelving, which it matched in all three runs, and it matched the workstation block in two runs, whereas Prompt-only placed the block 0.86~m from its reference center on average. On the remaining elements the conditions behaved alike, matching the workbench in every run and each desk row in two of three runs. The gain therefore rests on two elements, and one of the north-shelving matches lies exactly at the cover threshold (Section~\ref{sec:metrics}); counted as placed, it would lower the Measure mean to 3.7 and widen its range to 2--5.

On the third question, the elements that both conditions found lay at similar distances from the reference. Over the common set of four elements, the median centroid error averaged 0.31~m (0.22--0.50) for Measure and 0.29~m (0.13--0.39) for Prompt-only, and the workbench lay 0.16~m from the reference in both. Part of this error comes from the plan dimensions, which both conditions take from the metadata and which put the north wall 0.14~m south of the reference wall. The north desk row stands against that wall, and in every run its center lies 0.04 to 0.09~m south of the reference center, so the offset explains a small part of its mean errors of 0.29 and 0.30~m, and the remainder lies along the wall.

The measured placement came at a higher but still small cost in API usage and computation, which excludes preparing the plan and room metadata and taking the photo. A Measure run cost USD~1.63 and took 3.8~min on average, against USD~0.22 and 1.9~min for Prompt-only (Table~\ref{tab:results}), and a mean Measure run plus the single dress-and-verify run of Section~\ref{sec:dressresults} comes to about USD~11.3 and 42~min.

\begin{table}[t]
  \caption{Outcome of each fixed reference element in the three runs (M matched, P placed, -- missed) and mean centroid error (m) over the runs that found it. $^\dagger$Common set used for the medians. $^\ddagger$As in Table~\ref{tab:results}; the median of the common-set ($^\dagger$) element means is 0.26~m for Prompt-only and 0.34~m for Measure.}
  \label{tab:perref}
  \centering\footnotesize
  \setlength{\tabcolsep}{4pt}
  \begin{tabular}{l c r c r}
    \toprule
    & \multicolumn{2}{c}{Prompt-only} & \multicolumn{2}{c}{Measure} \\
    \cmidrule(lr){2-3}\cmidrule(lr){4-5}
    Reference element & runs & error & runs & error \\
    \midrule
    Workbench$^\dagger$ & M M M & 0.16 & M M M & 0.16 \\
    Tall shelving unit & P P P & 0.44 & P P -- & 0.65 \\
    South desk row$^\dagger$ & M M P & 0.23 & M P M & 0.37 \\
    North desk row$^\dagger$ & M P M & 0.29 & M P M & 0.30 \\
    North shelving & -- -- -- & & M M M & 0.31 \\
    West shelving & -- -- -- & & -- -- -- & \\
    Workstation block$^\dagger$ & P P P & 0.86 & M M P & 0.52 \\
    \midrule
    Mean of per-run medians$^\ddagger$ & & 0.29 & & 0.31 \\
    \bottomrule
  \end{tabular}
\end{table}

Two elements remained out of reach for both conditions, and both show limits of a single handheld photo. No run found the west shelving, which runs along the west wall into the north-west corner, about 5 to 7.5~m from the camera, and is mostly hidden behind two large displays on mobile stands. No run matched the tall shelving unit on the east wall either. The photo shows a cream cabinet standing in front of the south end of that unit, and the boxes of both conditions either covered less than half of the unit or had centers more than 0.5~m from its center.

All six placement models pass the IFC4 schema check and the IDS of Section~\ref{sec:levels}, and the rule checks found few geometric faults. The clash rule flags one box in two of the three Measure runs, each a storage box that overlaps the east-wall column, and three boxes in one Prompt-only run, where the plausibility rule also flags a monitor stand.

\subsection{Dressing and verification}
\label{sec:dressresults}
\begin{figure}[t]
  \centering
  \includegraphics[width=0.72\columnwidth]{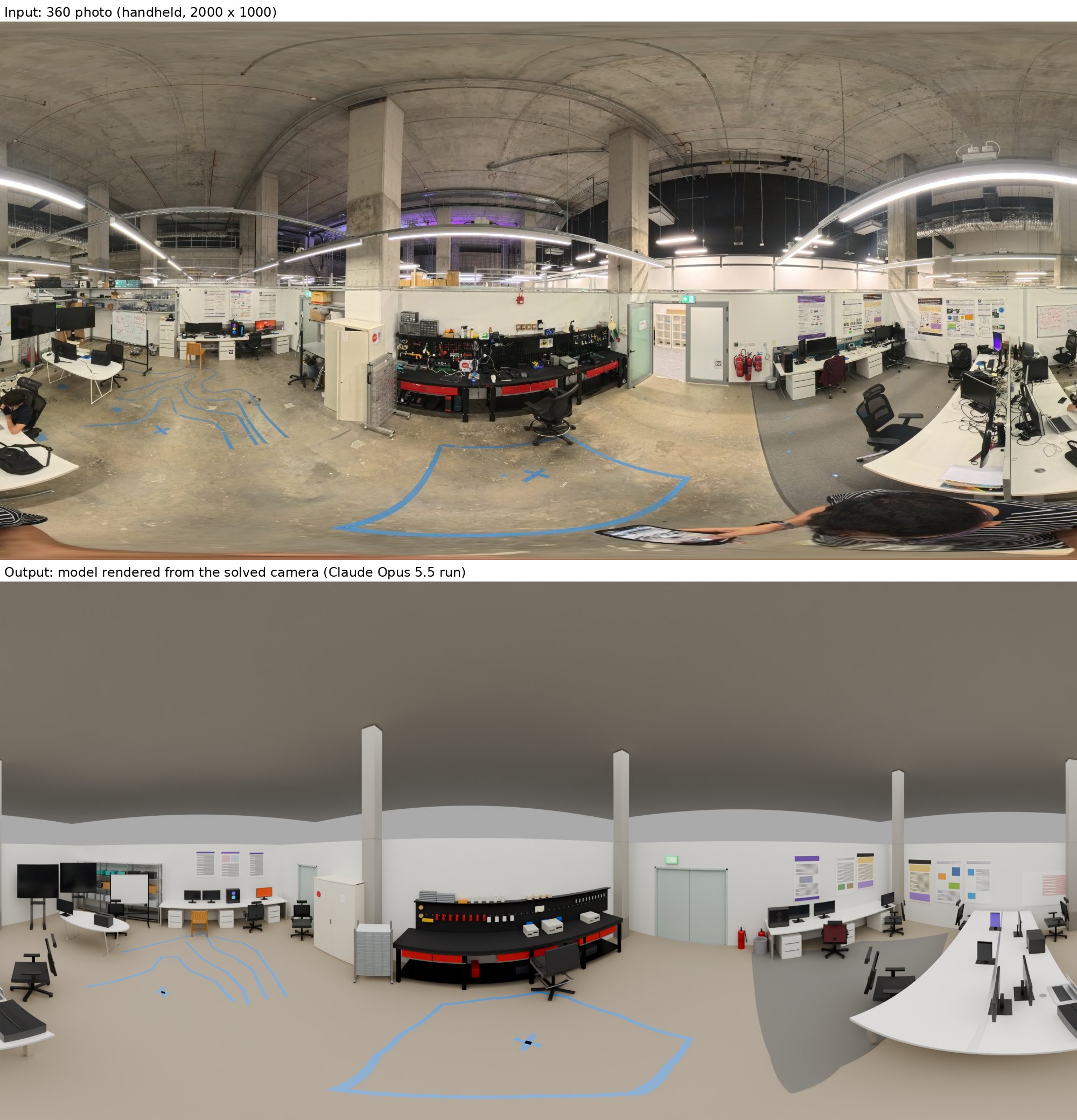}
  \caption{The dressed model resembles the photo from the camera position, judged by eye. The input 360\textdegree{} photograph (top) and the final presentation model after Stage~2 (Dress and Verify), rendered from the estimated camera pose (bottom), are both shown in equirectangular projection for a qualitative comparison.}
  \label{fig:photo_vs_model}
\end{figure}

The dress-and-verify stage ran end to end on the first Measure run and turned its seven boxes into a presentation model that resembles the photograph (Figs.~\ref{fig:before_after} and~\ref{fig:photo_vs_model}). It left all seven Measure boxes unchanged, so the placement scores of the first Measure run (Fig.~\ref{fig:plan}) also hold for the dressed model. The agent added 40 detail objects, including chairs and a stool, monitors, computer towers, a laptop, a mobile whiteboard, two display stands, a drawer rack, a pegboard, and two fire extinguishers. It used 21 object types and declared 11 wall items and 11 floor markings. The run made 26 model calls, cost USD~9.65, and took 38~min, of which about 15~min went to its four render checks on the two-core machine. Its last render check reported no hard physics issues, and the dressed model passes the IFC4 schema check and the IDS. No appearance measure was taken, so the resemblance to the photograph is judged by eye.

\section{Discussion}
\label{sec:discussion}

\subsection{Findings}
This paper set out to answer two questions. The first is whether a VLM that answers only through deterministic geometry tools can turn a floor plan with room dimensions and one 360\textdegree{} photo from a consumer camera into an IFC model of the room that comes close to an expert's model. The second is whether this constraint improves how much of the furniture the VLM finds and how closely it places it, compared with the same VLM writing every coordinate in one answer. On one room, one photo, and one expert reference, the answer to the first is yes at the LOD~200 level of approximation, with modest absolute agreement (Section~\ref{sec:limits}), and to the second partly yes, within the limits of agreement with that expert's model; accuracy against a survey remains to be shown.

The evidence follows the three contributions. In every Measure run the workflow took the plan and the handheld photo to an IFC4 model that passed the schema and IDS checks, and on the first of those runs Stage~2 dressed the model into a presentation model without moving a measured box; every decision in both stages was logged. The Measure runs reached a footprint IoU of 0.53 against 0.44 and matched 4.0 of the seven fixed reference elements per run against 2.3 for Prompt-only, at about USD~1.63 and 4~min per run. The elements that both conditions found lay at similar distances from the reference (median 0.31 and 0.29~m), so the gain came from elements that the single answer omitted or put far from their place, such as the north shelving and the workstation block.

The evaluation protocol of the third contribution shows why the three measures belong together. Footprint agreement and element outcomes register the gain, whereas centroid error on the common set alone would show none, since it covers only the elements both conditions found. A baseline must also receive the same modeling instructions before its errors can be attributed to geometry. Because the furniture layout changed between the photo and the reference, a digital twin should store movable elements with a capture date and report their agreement separately from that of installed furniture.

\subsection{Limitations}
\label{sec:limits}
The evidence comes from one room, one photograph, one model, and three runs per condition, on the space where the prompts and tools were developed, so it does not extend to other room types, multi-panorama capture, or unseen spaces. The three runs vary only the model's sampling and reuse the same photo, camera position, and lighting, so their ranges show model variability alone; variability from capture, which is likely larger in practice, is untested. All runs use one proprietary model, and the code, prompts, and execution traces are not publicly available but may be made available by the authors upon reasonable request.

The reference is an expert-authored model, and the furniture layout differs between the reference and the photograph. Consequently, the evaluation is limited to seven fixed furniture elements. The reference room dimensions also differ from those supplied with the floor plan by up to 0.14~m. Agreement with the reference may therefore reflect shared modeling errors, while discrepancies may arise from inaccuracies in the reference rather than the reconstruction. The Measure runs achieved a mean furniture-footprint IoU of 0.53 and did not match two of the seven fixed reference elements. The median centroid disagreement was approximately 0.3~m for the common set of four elements. These results quantify agreement with the expert-authored reference rather than absolute geometric accuracy. The accuracy and suitability of the reconstructed models for specific facility-management applications therefore remain to be established. The per-element uncertainty estimate is not calibrated either. On the Measure runs, the centroid error exceeded the reported value for 16 of the 17 element-run pairs that received boxes, with a median ratio of about 1.5.

Measure differs from Prompt-only in two ways at once, since it computes positions with tools and also views the panorama many times through zoomed windows; an agent with the same viewer that types its own coordinates would separate the two effects. Panoramic perception networks such as DeepPanoContext~\cite{zhang2021deeppanocontext} and PanoContext-Former~\cite{dong2024panocontextformer} were not compared, because they predict boxes for a fixed set of categories at a learned scale and produce no IFC model. Three runs per condition show the spread of the results but support no test of significance. Finally, the dress-and-verify stage was run once, leaves the LOD~200 layer unchanged, and has no appearance measure, so the results show that it runs and checks its own output, but they show neither that it improves model quality nor that the type library helps.

\subsection{Failure modes}
The runs exercised three of the failure classes anticipated in the design, pose ambiguity, the nadir region, and occlusion, and touched a fourth, clutter merged into one instance. Pose ambiguity arises because a rectangular room looks similar after a half turn. In the Measure runs the doors and columns broke the symmetry, and all six runs of both conditions gave headings between 332 and 338\textdegree, so no run turned the room. In the nadir region of a handheld shot the photographer occupies the bottom of the image (Fig.~\ref{fig:outlines}), and floor points read near it are unreliable, so the tool reports the distance of every floor point (Section~\ref{sec:placement}). Occlusion hid the west shelving behind two large displays on mobile stands in every run. Clutter merged into one instance is a design choice at LOD~200, where the grouping rule decides it. Non-Manhattan rooms and plan misreads were not exercised, and the glass door leaves and plain walls of the development space were not tested as separate conditions.

\subsection{Resolution limits}
At $2000\times1000$ pixels, one pixel spans 0.18\textdegree, or 1.6~cm at 5~m and 3~cm at 10~m. Screen depth is the hardest dimension to read. At the north desks, about 5.6~m from the camera, a 0.1~m extent across the line of sight spans about six pixels, and the depth of a monitor facing the camera only one or two, so the size and facing of screens are read with low confidence. A tripod capture at $5760\times2880$ would cut the pixel angle by a factor of about three. The per-element uncertainty estimate grows with distance so that users of the model can see where it is weak, although the estimate is not yet calibrated (Section~\ref{sec:limits}).

\section{Conclusion and Future Work}
\label{sec:future}

This paper presents Pano2BIM, a geometry-constrained vision-language agent workflow for generating as-built BIM models from a floor plan with room dimensions and a single 360\textdegree{} photograph. The VLM interprets visual information and identifies objects, while deterministic tools estimate camera pose, compute geometric placement, check spatial constraints, and generate IFC4 models with element-level provenance. In a preliminary evaluation on one laboratory with three runs per condition, Stage~1 (Measure) reached a footprint intersection over union (IoU) of 0.53 against an expert-authored reference and matched 4.0 of its seven fixed elements on average, against 0.44 and 2.3 for the single-call Prompt-only baseline with the same modeling instructions. On the elements that both found, the median centroid errors were similar (0.31 and 0.29~m), so the Measure workflow found and covered more of the furniture. Stage~2 (Dress and Verify) added 40 detail objects to the presentation model, and its final render check reported no hard physics issues. With a measure stage of about 4~min and USD~1.63 per run (about 42~min and USD~11 with the dress-and-verify stage), these preliminary findings point to rapid, low-cost generation of as-built BIM models, a key need for digital twins that reflect the buildings they represent.

Future work will focus on independently validating geometric accuracy, generalizability, and model-updating capabilities. Surveys of the development space and additional campus spaces, made at the time of capture, will provide ground truth for element positions, dimensions, orientations, and uncertainty estimates. Experiments on unseen spaces, with frozen prompts and tools, will assess generalization, and an agent that views the panorama like Measure but types its own coordinates will separate the effect of measured placement from that of closer viewing. Further ablations will investigate the contributions of rule-based checks, render-and-compare refinement, alternative perception methods, and multiple panoramic views. Additional studies will measure appearance, evaluate object-library transfer across environments, and use repeated image captures to identify physical changes and update BIM models over time. The code, tool definitions, prompts, experimental configurations, evaluation scripts, and execution traces may be made available by the authors upon reasonable request.

\section*{Acknowledgment}
This research was partially supported by the Center for Sand Hazards and Opportunities for Resilience, Energy, and Sustainability (SHORES), funded by Tamkeen under the NYUAD Research Institute Award CG013.

\bibliographystyle{IEEEtran}
\bibliography{refs}

\end{document}